\documentclass[twocolumn,twoside]{IEEEtran}
\usepackage{amsmath,amssymb,amsthm,epsfig,color,empheq,graphicx}
\usepackage{enumerate,url,algpseudocode,algorithm,wasysym,epstopdf}
\usepackage[table]{xcolor}
\usepackage{accents}
\usepackage[utf8]{inputenc}

\usepackage{subfig}

\def\h{{\mathbf h}}

\def\r{{\mathbf r}}
\def\s{{\mathbf s}}

\def\v{{\mathbf v}}

\def\z{{\mathbf z}}

\theoremstyle{remark}

\begin{document}
\title{Real-time Power System State Estimation and Forecasting via Deep Neural Networks}

\author{Liang Zhang,~\IEEEmembership{Student Member,~IEEE,}
	Gang Wang,~\IEEEmembership{Member,~IEEE,}
	and Georgios B. Giannakis,~\IEEEmembership{Fellow,~IEEE}

\vspace*{-0.7em}
	
\thanks{ }}

\maketitle

\begin{abstract}
Contemporary power grids are being challenged by rapid voltage fluctuations that are caused by large-scale deployment of renewable generation, electric vehicles, and demand response programs. In this context, monitoring the grid's operating conditions in real time becomes increasingly critical. With the emergent large scale and nonconvexity however, the existing power system state estimation (PSSE) schemes become computationally expensive or yield suboptimal performance. To bypass these hurdles, this paper advocates deep neural networks (DNNs) for real-time power system monitoring. By unrolling an iterative physics-based prox-linear solver, a novel model-specific DNN is developed for real-time PSSE with affordable training and  minimal tuning effort. To further enable system awareness even ahead of the time horizon, as well as to endow the DNN-based estimator with resilience, deep recurrent neural networks (RNNs) are also pursued for power system state forecasting. Deep RNNs leverage the long-term nonlinear dependencies present in the historical voltage time series to enable forecasting, and they are easy to implement. Numerical tests showcase improved performance of the proposed DNN-based estimation and forecasting approaches compared with existing alternatives. In real load data experiments on the IEEE $118$-bus benchmark system, the novel model-specific DNN-based PSSE scheme outperforms nearly by an order-of-magnitude the
competing alternatives, including the widely adopted Gauss-Newton PSSE solver.
\end{abstract}


\section{Introduction}\label{sec:intro}

Recognized as the most significant engineering achievement of the twentieth century, the North American power grid is a complex cyber-physical system with transmission and distribution infrastructure delivering electricity from generators to consumers. Due to the growing deployment of distributed renewable generators, electric vehicles, and demand response, contemporary power grids are facing major challenges related to unprecedented levels of load peaks and voltage fluctuations. In this context, real-time monitoring of the smart power grids becomes increasingly critical, not only for detection of system instabilities and protection~\cite{chap2017kwzg, dehghanpour2018survey, SPM2013, zhang2017going}, but also for energy management~\cite{SPM2013,liang16scalable, VZ}.

Given the grid parameters along with a set of measurements provided by the supervisory control and data acquisition (SCADA) system, PSSE aims to retrieve the unknown system state, that is, the complex voltages at all buses~\cite{chap2017kwzg}. Commonly used state estimators include the weighted least-squares (WLS) and least-absolute-value (LAV) ones, derived based on the (weighted) $\ell_1$- or $\ell_2$-loss criteria. To tackle the resultant nonconvex  optimization, 
various solvers have been proposed; see e.g. \cite{WollenbergBook,chap2017kwzg,dehghanpour2018survey, tsp2017wzgs}. However, those optimization-oriented PSSE schemes either require many iterations or are computationally intensive, and they are further challenged by growing dynamics and system size. These considerations motivate innovative approaches for real-time large-scale PSSE. 

Toward that end, PSSE using plain feed-forward neural networks (FNNs) was studied in \cite{barbeiro2014state,2018zfs}. Once trained off-line using historical data and/or simulated samples, FNNs can be implemented for real-time PSSE, as the inference 
entails only a few matrix-vector multiplications. Related approaches using FNNs that `learn-to-optimize' emerge in wireless communications \cite{sun2018learning},
\cite{2017sdw}, and outage detection \cite{zhao2017learning}. Unfortunately, past `plain-vanilla' FNN-based schemes not only suffer from `vanishing' or `exploding' gradients for deep nets, but are also model-agnostic, in the sense that they ignore the underlying physics of power grids, therefore lacking physical interpretation and yielding suboptimal performance. To devise NNs in a disciplined manner, recent proposals in computer vision \cite{gregor2010learning,sun2016deep} 
constructed deep (D) NNs by unfolding iterative solvers tailored to model-based optimization problems.  
 
In this work, we will pursue a hybrid approach that combines data with basic physical constraints, to develop model-specific DNNs for PSSE. On the other hand, PSSE by itself is insufficient for fully monitoring the system since it does not account for variations in the states (namely, system dynamics) \cite{chap2017kwzg}. In addition, PSSE works (well) only if there are enough measurements achieving system observability, and the grid topology along with the link parameters are precisely known. To address these challenges, power system state forecasting to aid PSSE~\cite{do2009forecasting, pesgm2014gwskgg, zhao2018robust, rosenthal2018ensemble} is well motivated. 
 
Power system state forecasting has so far been pursued via (extended) Kalman filtering and moving horizon approaches in e.g., \cite{debs1970dynamic},  \cite{da1983state,hassanzadeh2012power}, and also through first-order vector auto-regressive (VAR) modeling \cite{hassanzadeh2016short}. Nonetheless, all the aforementioned state predictors, assume \emph{linear} dynamics; yet in practice, the dependence of the current state on previous (estimated) one(s) is nonlinear and cannot be accurately characterized. To render \emph{nonlinear} estimators tractable, FNN-based state prediction has been advocated with the transition mapping modeled by a single-hidden-layer FNN \cite{do2009forecasting, do2009forecasting2}. Unfortunately, the number of FNN parameters grows linearly with the length of the input sequences, discouraging FNNs from capturing long-term dependencies in voltage time series. 

Our contribution towards real-time and accurate monitoring of the smart power grids  is two-fold. First, we advocate model-specific DNNs for PSSE, by unrolling a recently proposed prox-linear SE solver \cite{2017wgc}. Toward this goal, we first develop a reduced-complexity prox-linear solver. Subsequently, we demonstrate how the unrolled prox-linear solver lends itself to building blocks (layers) in contemporary DNNs. In contrast to `plain-vanilla' FNNs, our novel prox-linear nets require minimal tuning efforts, and come naturally with `skip-connections,' a critical element in modern successful DNN architectures (see \cite{he2016deep}) that enables efficient training. Moreover, to enhance system observability as well as enable system awareness ahead of time, we advocate power system state forecasting via deep recurrent NNs (RNNs). Deep RNNs enjoy a fixed number of parameters even with variable-length input sequences, and they are able to capture complex nonlinear dependencies present in time series data. Finally, we present numerical tests using real load data on the IEEE $57$- and $118$-bus benchmark systems to corroborate the merits of the developed deep prox-linear nets and RNNs relative to existing alternatives. 

The remainder of this paper is organized as follows. Section~\ref{sec: PSSE} outlines the basics of PSSE. Section~\ref{sec:PLN} introduces our novel reduced-complexity prox-linear solver for PSSE, and advocates the prox-linear net. Section~\ref{sec:RNN} deals with deep RNN for state forecasting, as well as shows how state forecasting can aid in turn DNN-based PSSE. Simulated tests are presented in Section~\ref{sec:simu}, while the paper is concluded with the research outlook in Section \ref{sec:conclusions}. 

\section{Least-absolute-value Estimation} \label{sec: PSSE}

Consider a power network consisting of $N$ buses that can be modeled as a graph $\mathcal{G}:=\{\mathcal{N},\mathcal{L}\}$, where $\mathcal{N}:=\{1,2,\ldots,N\}$ comprises all buses, and $\mathcal{L}:=\{(n,n')\}\in\mathcal{N}\times \mathcal{N}$ collects all lines. For each bus $n\in\mathcal{N}$, let $V_n := v_n^r + jv_n^i$ denote its corresponding complex voltage, and $P_n $ $ (Q_n)$ denote the active (reactive) power injection. For each line $(n,n') \in \mathcal{L}$, let $P_{nn'}^f $ $(Q_{nn'}^f)$ denote the active (reactive) power flow seen at the `forwarding' end, and $P_{nn'}^t$ ($Q_{nn'}^t$) denote the active (reactive) power flow at the `terminal' end. To perform power system state estimation or forecasting, 
suppose $M_t$ system variables are measured at time $t$. For a compact representation, let $\z_t:=$ $[\{|V_{n,t}|^2\}_{n\in \mathcal{N}_t^o},\, \{P_{n,t}\}_{n\in \mathcal{N}_t^o},\, \{Q_{n,t}\}_{n\in \mathcal{N}_t^o},   \{P_{nn',t}^f\}_{(n,n')\in \mathcal{E}_t^o},$ $\{Q_{nn',t}^f\}_{(n,n')\in \mathcal{E}_t^o}, \, \{P_{nn',t}^t\}_{(n,n')\in \mathcal{E}_t^o}, \, \{Q_{nn',t}^t\}_{(n,n')\in \mathcal{E}_t^o}]^\top$ be the data vector that collects all measured quantities at time $t$, where sets $\mathcal{N}_t^o$ and $\mathcal{E}_t^o$ signify  the locations where the corresponding nodal and
line quantities are measured.

Per time slot $t$, PSSE aims to recover the system state vector $\v_t:= [v_1^r, v_1^i, \ldots,v_N^r, v_N^i]^\top  \in \mathbb{R}^{2N}$ (expanded in the rectangular coordinates) from generally noisy data $\z_t$. For brevity, the subscript $t$ of $\z_t$ and $\v_t$ will be omitted when discussing PSSE in Sections~\ref{sec: PSSE} and~\ref{sec:PLN}. Concretely, the PSSE task can be posed as follows. 

Given $M$ measurements $\{ z_m\}_{m=1}^M$ and the corresponding matrices $\{\mathbf{H}_m \in\mathbb{R}^{2N\times 2N}\}_{m=1}^M$ obeying the physical model 
\begin{equation}\label{eq:data}
	z_m=\mathbf{v}^\top\mathbf{H}_m\mathbf{v}+\epsilon_m, \qquad \forall m=1, \ldots, M
\end{equation}
our goal is to recover $\mathbf{v}\in\mathbb{R}^{2N}$, where $\{\epsilon_m\}$ account for the measurement noise and modeling inaccuracies. 

Adopting the LAV error criterion that is known to be robust to outliers, the ensuing LAV estimate is sought (see e.g., \cite{WollenbergBook})
\begin{equation}\label{eq:lav}
\hat{\v} :=	\arg\min_{\mathbf{v}\in\mathbb{R}^{2N}} \frac{1}{M} \sum_{m=1}^M \left| 
	z_{m} -\mathbf{v}^\top{\mathbf{H}_m}\mathbf{v}\right |
\end{equation}
for which various solvers have been developed \cite{1991fast,jabr2004,chap2017kwzg}. In particular, the recent prox-linear solver developed in \cite{2017wgc} has well documented merits, including provably fast (locally quadratic) convergence, as well as efficiency in coping with the  non-smoothness and nonconvexity in \eqref{eq:lav}.
Specifically, the prox-linear solver starts with some initial vector $\mathbf{v}_0$, and iteratively minimizes the regularized and `locally linearized' (relaxed) cost in \eqref{eq:lav}, to obtain iteratively (see also \cite{1996gn, tcns2018wzgs})
\begin{equation}\label{eq:proxl}
	{\mathbf{v}}_{i+1}=\arg\underset{\mathbf{v}\in \mathbb{R}^{2N}}{\min}~ 
	\left\| \mathbf{z} - \mathbf{J}_i(2\mathbf{v}-\mathbf{v}_i) \right\|_1+\frac{M}{2\mu_i}\left\|\mathbf{v}-\mathbf{v}_i\right\|_2^2
\end{equation} 
where $i\in \mathbb{N}$ is the iteration index, 
$\mathbf{J}_i:=[\mathbf{v}_i^\top\mathbf{H}_m ]_{1\le m\le M}$ is an $M\times N$ matrix whose $m$-th row is $\mathbf{v}_i^\top\mathbf{H}_m$, and $\{\mu_i>0\}$ is a pre-selected step-size sequence.  

It is clear that the per-iteration subproblem \eqref{eq:proxl} is a convex quadratic program, which can be solved by means of standard convex programming methods. One possible iterative solver of \eqref{eq:proxl} is based on the alternating direction method of multipliers (ADMM). Such an ADMM-based inner loop turns out to entail  $2M+2N$ auxiliary variables, thus requiring the update of $2M+4N$ variables per iteration~\cite{2017wgc}. 

Aiming at a reduced-complexity solver, in the next section we will first recast \eqref{eq:proxl} in a Lasso-type form, and subsequently unroll the resultant  double-loop prox-linear iterations~\eqref{eq:proxl} that constitute the key blocks of our DNN-based PSSE solver.

\begin{figure*}[t]
	\centering
	\includegraphics[scale=0.42]{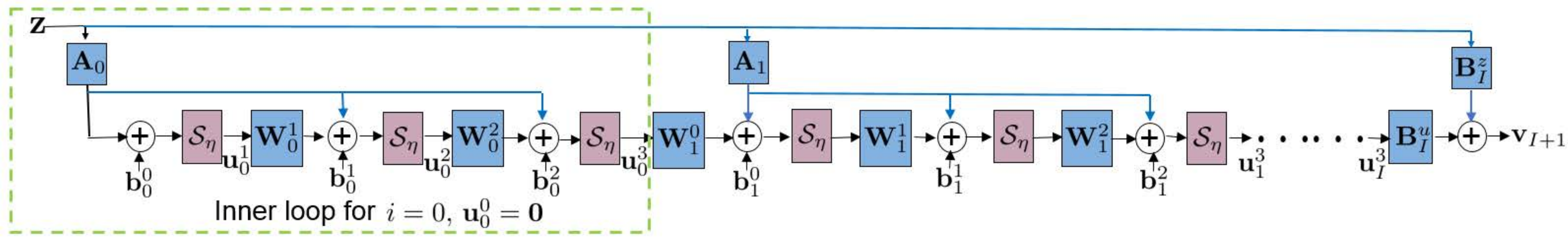}
	\caption{Prox-linear net with $K=3$ blocks.}
	\label{fig:plnet}
\end{figure*}

\begin{figure*}[t]
	\centering
	\includegraphics[scale=0.43]{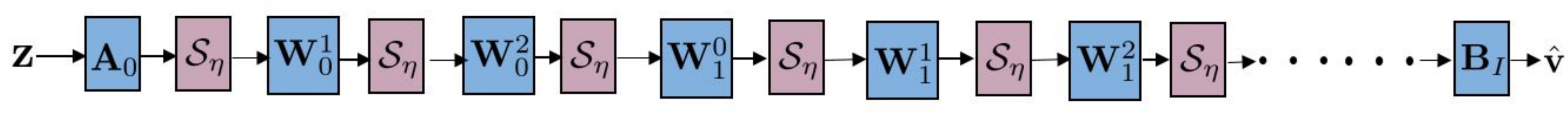}
		\vspace{-.2em}
	\caption{Plain-vanilla FNN which has the same per-layer number of hidden units as the prox-linear net.}
	\label{fig:pvnet}
\end{figure*}

\section{The Prox-linear Net}\label{sec:PLN}
In this section, we will develop a DNN-based scheme to approximate the solution of~\eqref{eq:lav} by unrolling the double-loop prox-linear iterations. Upon defining the vector variable $\mathbf{u}_i:= \mathbf{J}_i(2\mathbf{v}-\mathbf{v}_i)-\mathbf{z}$, and plugging $\mathbf{u}_i$ into \eqref{eq:proxl}, we arrive at 
\begin{equation}\label{eq:solveproxliner}
	{\mathbf{u}}_i^*=\arg\underset{\mathbf{u}_i\in\mathbb{R}^M}{\min}~ 
	\left\|\mathbf{u}_i\right\|_1+\frac{M}{4\mu_i}\left\|\mathbf{B}_i(\mathbf{u}_i +\mathbf{z})- \mathbf{v}_i\right\|_2^2
\end{equation}     
where $\mathbf{B}_i\in \mathbb{R}^{2N\times M}$ denotes the pseudo inverse of $\mathbf{J}_i$ that satisfies $\mathbf{B}_i\mathbf{J}_i = \mathbf{I}_{2N\times 2N}$. The pseudo inverse $\mathbf{B}_i$ exists because PSSE requires $M \geq 2N$ to guarantee system observability. 

Once the inner optimum variable $\mathbf{u}_i^*$ is found, the next outer-loop iterate $\mathbf{v}_{i+1}$ can be readily obtained as 
\begin{equation} \label{eq:vt1}
\mathbf{v}_{i+1}=  [\mathbf{B}_i(\mathbf{u}_i^* +\mathbf{z})+ \mathbf{v}_i]/2
\end{equation} 
following the definition of $\mathbf{u}_i$. Interestingly, \eqref{eq:solveproxliner}~now reduces to a Lasso problem \cite{proximal}, for which various celebrated solvers have been put forward, including e.g., the iterative shrinkage and thresholding algorithm (ISTA)~\cite{proximal}. 

Specifically, with $k$ denoting the iteration index of the inner-loop, the ISTA for \eqref{eq:solveproxliner} proceeds across iterations $k= 0, 1, \ldots$ along the following recursion
\begin{align} \label{eq:proxgd}
	{\mathbf{u}}_i^{k+1} &= \mathcal{S}_{\eta}\!\left( \mathbf{u}_i^k - \frac{\eta M}{2\mu_i} \mathbf{B}_i^\top\left[\mathbf{B}_i\!\left(\mathbf{u}_i^k+ \mathbf{z}\right) -\mathbf{v}_i \right]\right)\\
	&= \mathcal{S}_{\eta}\!\left(\mathbf{W}_i^k \mathbf{u}_i^k + \mathbf{A}_i\mathbf{z} + \mathbf{b}_i^k\right) \nonumber   
\end{align} 
where $\eta>0$ is a fixed step size with coefficients
\begin{subequations}
	\label{eq:cof}
	\begin{align}
		\mathbf{W}_i^k &:= \mathbf{I} - \frac{\eta  M}{2\mu_i} \mathbf{B}_i^\top\mathbf{B}_i,\quad \forall k\in \mathbb{N} \\
		\mathbf{A}_i &:= - \frac{\eta  M}{2\mu_i} \mathbf{B}_i^\top\mathbf{B}_i\\ 
		\mathbf{b}_i^k &:=  \frac{\eta M}{2\mu_i} \mathbf{B}_i^\top\mathbf{v}_i,\quad \forall k\in \mathbb{N} 
	\end{align}
\end{subequations}
and
$\mathcal{S}_{\eta}(\cdot)$ is the so-termed soft thresholding operator 
\begin{equation}\label{eq:softt}
	\mathcal{S}_{\eta}(x) :=\left\{ \begin{array}{ll}
		x - \eta,		&~	 x > \eta  \\
		0,						&~	-\eta \leq x\leq \eta\\
		x + \eta,	&~	  x < -\eta\\
	\end{array}\right.
\end{equation}
understood entry-wise when applied to a vector. With regards to initialization, one can set $\mathbf{u}_0^0=\bf{0}$ without loss of generality,
and $\mathbf{u}_{i}^0 = \mathbf{u}_{i-1}^*$ for $i\ge 1$.

\begin{algorithm}[h!]
	\caption{Reduced-complexity prox-linear solver.}\label{alg:pla}
	\begin{algorithmic}[1]
		\renewcommand{\algorithmicrequire}{\textbf{Input:}}
		\renewcommand{\algorithmicensure}{\textbf{Output:}}
		\Require Data $\{(z_m, \mathbf{H}_m)\}_{m=1}^M$,~step sizes $\{\mu_i\},\, \eta $, and initialization $\mathbf{v}_0= \mathbf 1$, $\mathbf{u}_0^0= \mathbf 0$.
		{\For{$i = 0,1, \ldots,I$}
			\State Evaluate $\mathbf{W}_i^k,\, \mathbf{A}_i$, and $\mathbf{b}_i^k$  according to \eqref{eq:cof}.
			\State Initialize $\mathbf{u}_i^0$.
			\For{$k = 0,1, \ldots,K$}
			\State Update $\mathbf{u}_i^{k+1}$ using \eqref{eq:proxgd}.
			\EndFor 
			\State 		 Update $\mathbf{v}_{i+1}$ using \eqref{eq:vt1}.
			\EndFor}
	\end{algorithmic}
\end{algorithm}  

The new prox-linear PSSE solver with reduced-complexity is summarized in Alg.~\ref{alg:pla}. With appropriate step sizes $\{\mu_i\}$ and $\eta$, the sequence $\{\mathbf{v}_i\}$ generated by Alg.~\ref{alg:pla} converges to a stationary point of \eqref{eq:lav}~\cite{2017wgc}. In practice, Alg.~\ref{alg:pla} often requires a large number $K$ of inner iterations to approximate the solution of \eqref{eq:solveproxliner}. Furthermore, the pseudo-inverse $\mathbf{B}_i$ has to be computed per outer-loop iteration. These challenges can limit its use in large-scale as well as in real-time applications. 

Instead of solving the optimization problem with a (large) number of iterations, recent proposals \cite{gregor2010learning,sun2016deep} advocated trainable DNNs constructed by unfolding those iterations, to obtain data-driven solutions. As demonstrated in \cite{gregor2010learning,wang2016learning,sun2016deep}, properly trained unrolled DNNs can achieve competitive performance even with a small number of layers. In the following, we will elaborate on how to unroll our Alg.~\ref{alg:pla} to construct DNNs for high-performance PSSE. 

Consider first unrolling the outer loop~\eqref{eq:proxl} up to, say the $(I+1)$-st iteration to obtain $\mathbf{v}_{I+1}$. Leveraging the recursion \eqref{eq:proxgd}, each inner loop iteration $i$ refines the initialization $\mathbf{u}_i^0= \mathbf{u}_{i-1}^K$ to yield after $K$ inner iterations $\mathbf{u}_{i}^K$. Such an unrolling leads to a $K(I+1)$-layer structured DNN. Suppose that the sequence $\{\mathbf{v}_i\}_{i=0}^{I+1}$ has converged, which means $\|\mathbf{v}_I- \mathbf{v}_{I+1}\|\le \epsilon$ for some $\epsilon>0$. It can then be deduced that $\mathbf{v}_{I+1}= \mathbf{B}_I^u\mathbf{u}_I^* + \mathbf{B}_I^z\mathbf{z}$ with $\mathbf{B}_I^u:= \mathbf{B}_I$ and $\mathbf{B}_I^z:= \mathbf{B}_I$. 

Our novel DNN architecture, that builds on the physics-based Alg. \ref{alg:pla}, is thus a hybrid combining plain-vanilla FNNs with the conventional iterative solver such as Alg. \ref{alg:pla}. We will henceforth term it `prox-linear net.' For illustration, the prox-linear net with $K=3$ is depicted in~Fig.~\ref{fig:plnet}. The first inner loop $i=0$ is highlighted in a dashed box, where $\mathbf{u}_0^1= \mathcal{S}_{\eta}( \mathbf{A}_0\mathbf{z} + \mathbf{b}_0)$ because $\mathbf{u}_0^0 = \mathbf{0}$. As with \cite{gregor2010learning,wang2016learning,sun2016deep}, our prox-linear net can be treated as a trainable regressor that predicts $\mathbf{v}$ from data $\mathbf{z}$, where the coefficients $\{\mathbf{b}_i^k\}_{0\le i \le I}^{1\le k\le 3}$, $\{\mathbf{A}_i\}_{i=0}^I$,  $\{\mathbf{W}_i^k\}_{0\le i \le I}^{1\le k\le 3}$,  $\mathbf{B}_I^u$, and $\mathbf{B}_I^z$ are typically untied to enhance  approximation capability and learning flexibility. Given historical and/or simulated measurement-voltage training pairs $\{(\mathbf{z}_s,\mathbf{v}_s) \}$, these coefficients can be learned end-to-end using backpropagation  \cite{backprop}, possibly employing the Huber loss \cite{book2011huber} to endow the state estimates with resilience to outliers. 

Relative to the conventional FNN in Fig.~\ref{fig:pvnet}, our proposed prox-linear net features: i) `skip-connections' (the bluish
lines in Fig.~\ref{fig:plnet}) that directly connect the input $\mathbf{z}$ to intermediate/output layers, and ii) a fixed number ($M$ in this case) of hidden neurons per layer. It has been shown both analytically and empirically that these skip-connections help avoid the so-termed `vanishing' and `exploding' gradient issues, therefore enabling successful and efficient training of DNNs~\cite{he2016deep,he2016identity}. The `skip-connections' is also a key enabler of the universal approximation capability of DNNs with a fixed number of hidden-neurons per-layer \cite{lin2018resnet}. 

The only hyper-parameters that must be tuned in our prox-linear net are $I$, $K$, and $\eta$, which are also tuning parameters required by the iterative optimization solver in Alg. \ref{alg:pla}. It is also worth pointing out that other than the soft-thresholding nonlinearity (a.k.a. activation function) used in Figs.~\ref{fig:plnet} and~\ref{fig:pvnet}, alternative functions such as the rectified linear unit (ReLU) (see e.g. \cite{relus2018wgc} for its good properties)  can be applied as well~\cite{gregor2010learning}. We have observed in our simulated tests that the prox-linear net with soft thresholding operators or ReLUs yield similar performance. To understand how different network architectures affect the performance, ReLU
activation functions are used by default unless otherwise stated. 

\begin{figure}[t]
	\centering
	\includegraphics[scale=0.48]{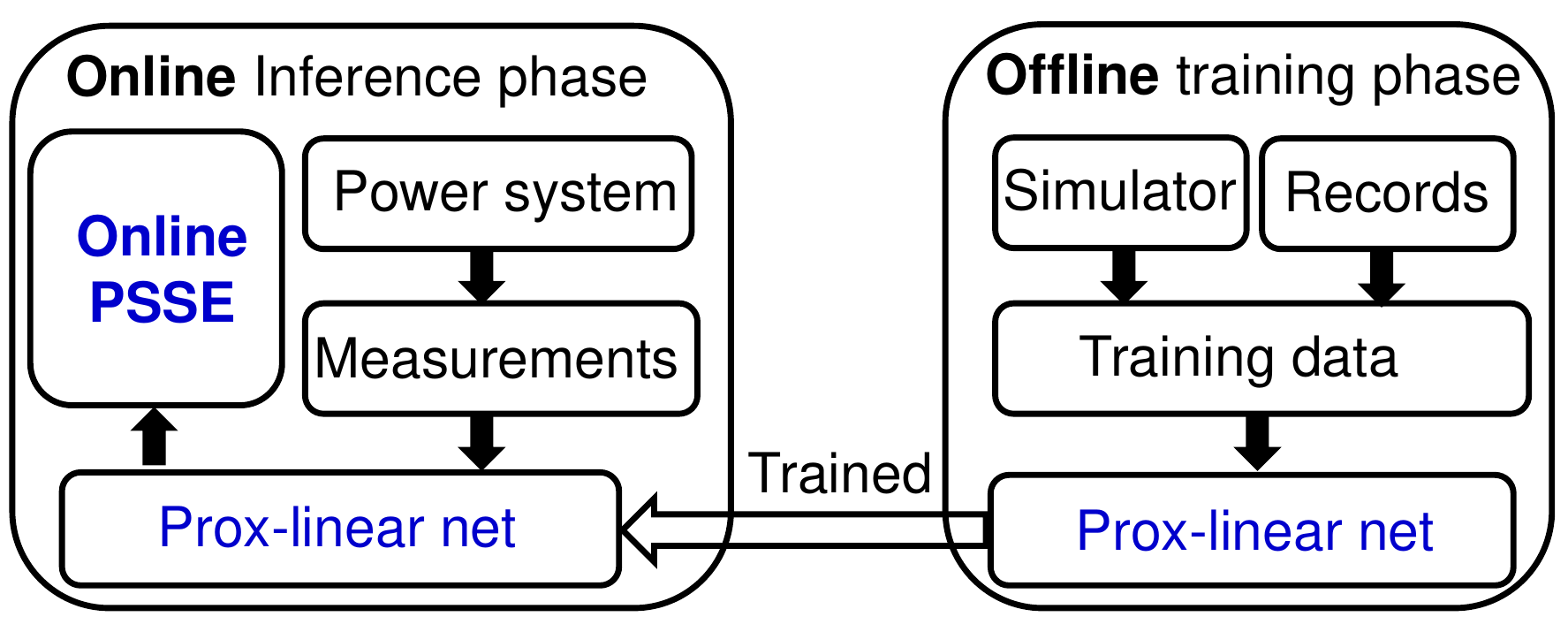}
	\vspace{.4em}
	\caption{Deep prox-linear net based real-time PSSE.}
	\label{fig:flowchart}
	\vspace{-.6em}
\end{figure}

The flow chart demonstrating the prox-linear net for real-time PSSE is depicted in Fig.~\ref{fig:flowchart}, where the real-time inference stage is described in the  
left rounded rectangular box, while the off-line training stage is on the left. Thanks to the wedding of the physics in \eqref{eq:data} with our DNN architecture design, the extensive numerical tests in Section \ref{sec:simu} will confirm an impressive boost in performance of our prox-linear nets relative to competing FNN and Gauss-Newton based PSSE approaches.

\section{Deep RNNs for State Forecasting} \label{sec:RNN}

Per time slot $t$, the PSSE scheme we developed in  Section~\ref{sec:PLN} estimates the state vector $\mathbf{v}_t \in \mathbb{R}^{2N}$ upon receiving measurements $\mathbf{z}_t$. Nevertheless, its performance is challenged when there are missing entries in $\mathbf{z}_t$, which is indeed common in a SCADA system due for example to meter and/or communication failures \cite{chap2017kwzg}. To enhance our novel PSSE scheme and obtain system awareness even ahead of time, we are prompted to pursue power system state forecasting, which for a single step amounts to predicting the next state $\mathbf{v}_{t+1}$ at time slot $t+1$ from the available time-series $\{{\mathbf{v}}_{\tau}\}_{\tau = 0}^t$~\cite{do2009forecasting}. Analytically, the estimation and prediction steps are as follows 
\begin{align}
	\mathbf{v}_{t+1} &= \boldsymbol{\phi}({\mathbf{v}}_t, {\mathbf{v}}_{t-1}, {\mathbf{v}}_{t-2}, \ldots, {\mathbf{v}}_{t-r +1}) +\boldsymbol{\xi}_t \label{eq:fv} \\ 
	\mathbf{z}_{t+1} &= \mathbf{h}_{t+1}(\mathbf{v}_{t+1}) + \boldsymbol{\epsilon}_{t+1}
	\label{eq:s2m} 
\end{align}
where  $\{\boldsymbol{\xi}_t,\boldsymbol{\epsilon}_{t+1}\}$ account for modeling inaccuracies; the tunable parameter $r\ge 1$ represents the number of lagged (present included) states used to predict $\v_{t+1}$; and the unknown (nonlinear) function $\boldsymbol{\phi}$ captures the state transition, while $\mathbf{h}_{t+1}(\cdot)$ is the measurement function that summarizes equations in \eqref{eq:data} at time slot $t+1$. To perform state forecasting, function $\boldsymbol{\phi}$ must be estimated or approximated -- a task that we will accomplish using RNN modeling, as we present next. 

RNNs are NN models designed to learn from correlated time series data. Relative to FNNs, RNNs are not only scalable to long-memory inputs (regressors) that entail sequences of large $r$, but are also capable of processing input sequences of variable length~\cite{Goodfellow-et-al-2016}. Given the input sequence $\{{\mathbf{v}}_\tau\}_{\tau = t-r+1}^{t}$, and an initial state $\s_{t-r}$, an RNN finds the \emph{hidden state}\footnote{Hidden state is an auxiliary set of vector variables not to be confused with the power system state $\mathbf{v}$ consisting of the nodal voltages as in \eqref{eq:data}.} vector sequence $\{\s_\tau\}_{\tau =  t-r +1}^t $ by repeating  
\begin{equation} \label{eq:ss}
	\mathbf{s}_\tau = f(\textcolor{black}{\mathbf{R}^{0}}\mathbf{v}_\tau + \mathbf{R}^{ss}\mathbf{s}_{\tau-1} + \mathbf{r}^0)
\end{equation}
where $f(\cdot)$ is a nonlinear activation function (e.g., a ReLU or sigmoid unit), understood entry-wise when applied to a vector, whereas the coefficient matrices $\mathbf{R}^{0}$, $\mathbf{R}^{ss}$, and the vector $\mathbf{r}^0$ contain time-invariant weights. 

\begin{figure}[t]
	\centering
	\includegraphics[scale=0.3]{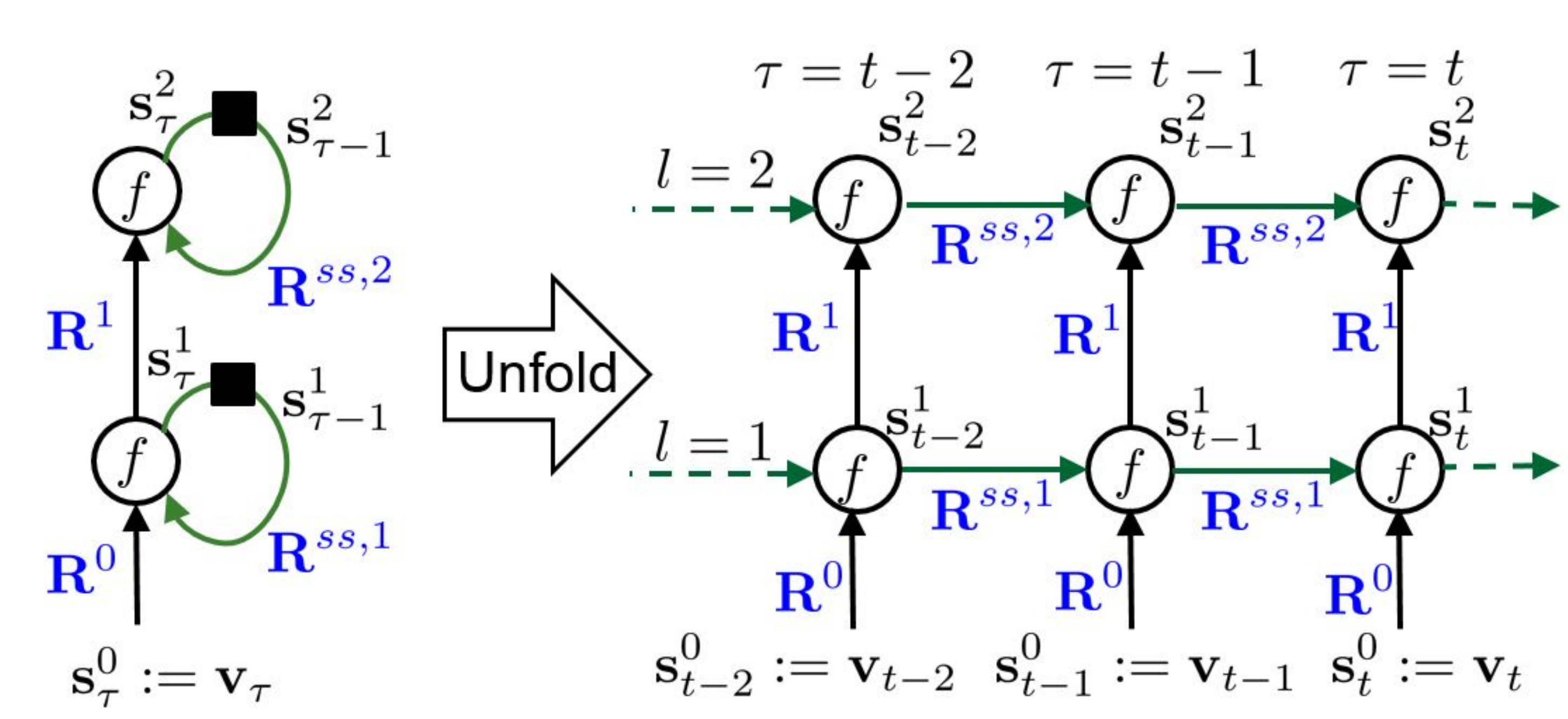}
	\caption{An unfolded deep RNN with no outputs.}
	\label{fig:rnn}
\end{figure}

Deep RNNs are RNNs of multiple ($\ge 3$) processing layers, which can learn compact representations of time series through hierarchical nonlinear transformations. The state-of-the-art in numerous sequence processing applications, including music prediction and machine translation~\cite{Goodfellow-et-al-2016}, has been significantly improved with deep RNN models. By stacking up multiple recurrent hidden layers (cf. \eqref{eq:ss}) one on top of another, deep RNNs can be constructed as follows~\cite{deepRNN}
\begin{equation} \label{eq:deepRNN}
	\mathbf{s}_\tau^l = f\big( \mathbf{R}^{l-1} \mathbf{s}_{\tau}^{l-1} + \mathbf{R}^{ss,l} \mathbf{s}_{\tau-1}^{l} + \r^{l-1}\big),\quad l \ge 1
\end{equation}
where $l$ is the layer index, $\mathbf{s}_\tau^l$ denotes the hidden state of the $l$-th layer at slot $\tau$ having $\mathbf{s}_\tau^0 := \v_\tau$, and $\{\mathbf{R}^{l},\mathbf{R}^{ss, l},\mathbf{r}^l \}$ collect all unknown weights. Fig.~\ref{fig:rnn} (left) depicts the computational graph representing \eqref{eq:deepRNN} for $l=2$, with the bias vectors ${\bf r}^l = {\bf 0}$, $\forall l$ for simplicity in depiction, and the black squares standing for single-step delay units. Unfolding the graph by breaking the loops and connecting the arrows to corresponding units of the next time slot, leads to a deep RNN in Fig.~\ref{fig:rnn} (right), whose rows represent layers, and columns denote time slots. 

The RNN output can come in various forms, including one output per time step, or, one output after several steps. The latter matches the $r$th-order nonlinear regression in~\eqref{eq:fv} when approximating $\boldsymbol{\phi}$ with a deep RNN. Concretely, the output of our deep RNN is given by
\begin{equation} \label{eq:output}
	\check{\mathbf{v}}_{t+1} = \mathbf{R}^{out}\mathbf{s}_t^l  + \mathbf{r}^{out}
\end{equation}
where $\check{\mathbf{v}}_{t+1}$ is the forecast of $\mathbf{v}_{t+1}$ at time $t$, and $(\mathbf{R}^{out},\mathbf{r}^{out})$ contain weights of the output layer. Given historical voltage time series, the weights $(\mathbf{R}^{out},\mathbf{r}^{out})$ and $\{\mathbf{R}^{l},\mathbf{R}^{ss,l},\mathbf{r}^l\}$ can be learned end-to-end using backpropagation \cite{Goodfellow-et-al-2016}. Invoking RNNs for state-space models, the class of nonlinear predictors discussed in~\cite{do2009forecasting} is considerably broadened here to have memory. As will be demonstrated through extensive numerical tests, the forecasting performance can be significantly improved through the use of deep RNNs. Although the focus here is on one-step state forecasting, it is worth stressing that our proposed approaches with minor modifications, can be generalized to predict the system states multiple steps ahead.

So far, we have elaborated on how RNNs enable flexible nonlinear predictors for power system state forecasting. To predict $\check{\mathbf{v}}_{t+1}$ at time slot $t$, the RNN in~\eqref{eq:deepRNN}  requires ground-truth voltages $\{{\mathbf{v}}_\tau\}_{\tau = t-r+1}^{t}$ (cf. \eqref{eq:fv}), which however, may not be available in practice. Instead we can use the estimated ones $\{\hat{\mathbf{v}}_\tau\}_{\tau = t-r+1}^{t}$ provided by our prox-linear net-based estimator in Section~\ref{sec:PLN}. In turn, the forecast $\check{\mathbf{v}}_{t+1}$ can be employed as a prior to aid PSSE at time slot $t+1$, by providing the so-termed virtual measurements  $\check{\z}_{t+1}:=\h_{t+1}(\check{\mathbf{v}}_{t+1})$ that can be readily accounted for in \eqref{eq:lav}. For example, when there are missing entries in $\mathbf{z}_{t+1}$, the obtained $\check{\z}_{t+1}$ can be used to improve the PSSE performance by imputing the missing values. 
 
\begin{figure}[t]
	\centering
	\includegraphics[scale=0.36]{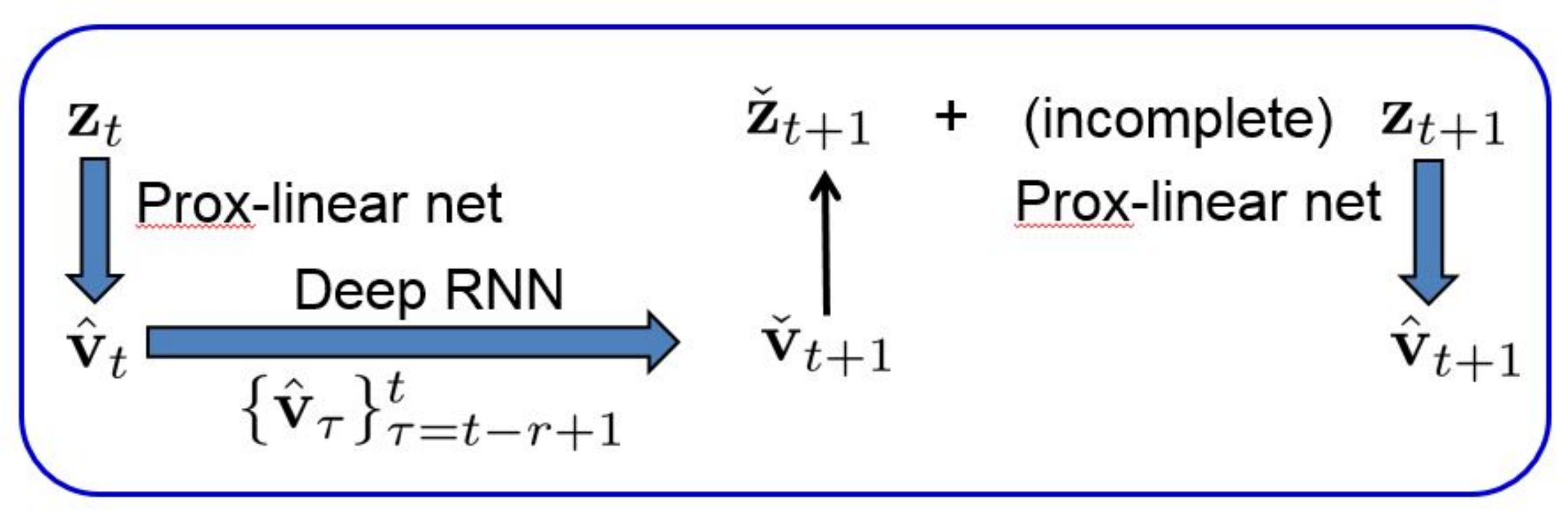}
	\vspace{.4em}
	\caption{DNN-based real-time power system monitoring.}
	\label{fig:monitor}
	\vspace{-.6em}
\end{figure}

Figure~\ref{fig:monitor} depicts the flow chart of the overall real-time power system monitoring scheme, consisting of deep prox-linear net-based PSSE and deep RNN-based state forecasting modules, that are implemented at time $t$ and $t+1$. Our novel  scheme is reminiscent of the predictor-corrector-type estimators emerging with dynamic state estimation problems using Kalman filters~\cite{zhao2018robust, chap2017kwzg}. Although beyond the scope of the present paper, it is worth remarking that the residuals $\mathbf{z}_{t+1} - \h_{t+1}(\hat{\mathbf{v}}_{t+1})$ along with $\mathbf{z}_{t+1} - \check{\z}_{t+1}$ can be  used to unveil erroneous data, as well as changes in the grid topology and the link parameters; see~\cite{do2009forecasting, GG2017graph} for an overview.

\section{Numerical Tests}\label{sec:simu}
Performance of our deep prox-linear net based PSSE, and deep RNN based state forecasting methods was evaluated using the IEEE $57$- and $118$-bus benchmark systems.  Real load data from the 2012 Global Energy Forecasting Competition (GEFC)\footnote{https://www.kaggle.com/c/global-energy-forecasting-competition-2012-load-forecasting/data.} were used to generate the training and testing datasets, where the load series were subsampled for size reduction by a factor of $5$ ($2$) for the IEEE $57$-bus ($118$-bus) system. Subsequently, the resultant load instances were normalized to match the scale of power demands in the simulated
system. The MATPOWER toolbox \cite{MATPOWER} was used to solve the AC power flow equations with the normalized load series as inputs, to obtain the ground-truth voltages $\{\mathbf{v}_\tau\}$, and produce
measurements $\{\mathbf{z}_\tau\}$ that comprise all forwarding-end active (reactive) power flows, as well as all voltage magnitudes. All NNs were trained using  `TensorFlow'~\cite{tensorflow2015-whitepaper} on an NVIDIA Titan X GPU with $12$GB RAM, with weights learned by the backpropagation based algorithm `Adam' (with starting learning rate $10^{-3}$) for $200$ epochs. To alleviate randomness in the obtained weights introduced by the
training algorithms, 
all NNs were trained and tested independently for $20$ times, with reported results averaged over $20$ runs. For reproducibility,
the `Python'-based implementation of our prox-linear net for PSSE of the $118$-bus system is publicly available at https://github.com/LiangZhangUMN/PSSE-via-DNNs.

\subsection{Prox-linear nets for PSSE}
To start, the prox-linear net based PSSE was tested, which estimates $\{\hat{\mathbf{v}}_\tau\}$ using $\{\mathbf{z}_\tau\}$. For both training and testing phases, all measurements $\{\mathbf{z}_\tau\}$ were corrupted by additive white Gaussian noise, where the  standard deviation for power flows and for voltage magnitudes was $0.02$ and $0.01$. The estimation performance of our prox-linear net was assessed in terms of the normalized root mean-square error (RMSE) $\|\hat{\mathbf{v}}-\mathbf{v}\|_2 / N$, where  $\mathbf{v}$ is the ground truth, and $\hat{\mathbf{v}}$  the estimate obtained by the prox-linear net. 
 
In particular, the prox-linear net was simulated with $T=2$ and $ K=3$. The `workhorse' Gauss-Newton method, a $6$-layer `plain-vanilla' FNN that has the same depth as our prox-linear net, and an $8$-layer `plain-vanilla' FNN that has roughly the same number of parameters as the prox-linear net, were simulated as baselines. The number of hidden units per layer in all NNs was kept equal to the dimension of the input, that is, $57 \times 2 = 114$ for the $57$-bus system and $118\times 2 = 236$ for the $118$-bus system. 

In the first experiment using the 57-bus system, a total of $7,676$ measurement-voltage $(\mathbf{z}_\tau, \mathbf{v}_\tau)$ pairs were generated, out of which the first $6,176$ pairs were used for training, and the rest were kept for testing. The average performance over $20$ trials, was evaluated in terms of the average RMSEs over the $1,500$ testing examples for the prox-linear net, Gauss-Newton, $6$-layer FNN, and $8$-layer FNN, are ${3.49\times10^{-4}}$, $3.2\times10^{-4}$, $6.35\times10^{-4}$, and $9.02\times10^{-4}$, respectively. These numbers showcase competitive performance of the prox-linear net. Interestingly, when the number of hidden layers of `plain-vanilla' FNNs increases from $6$ to $8$,  the performance degrades due partly to the difficulty in training the $8$-layer FNN. 

\begin{figure}[t]
	\centering
	\includegraphics[scale=0.58]{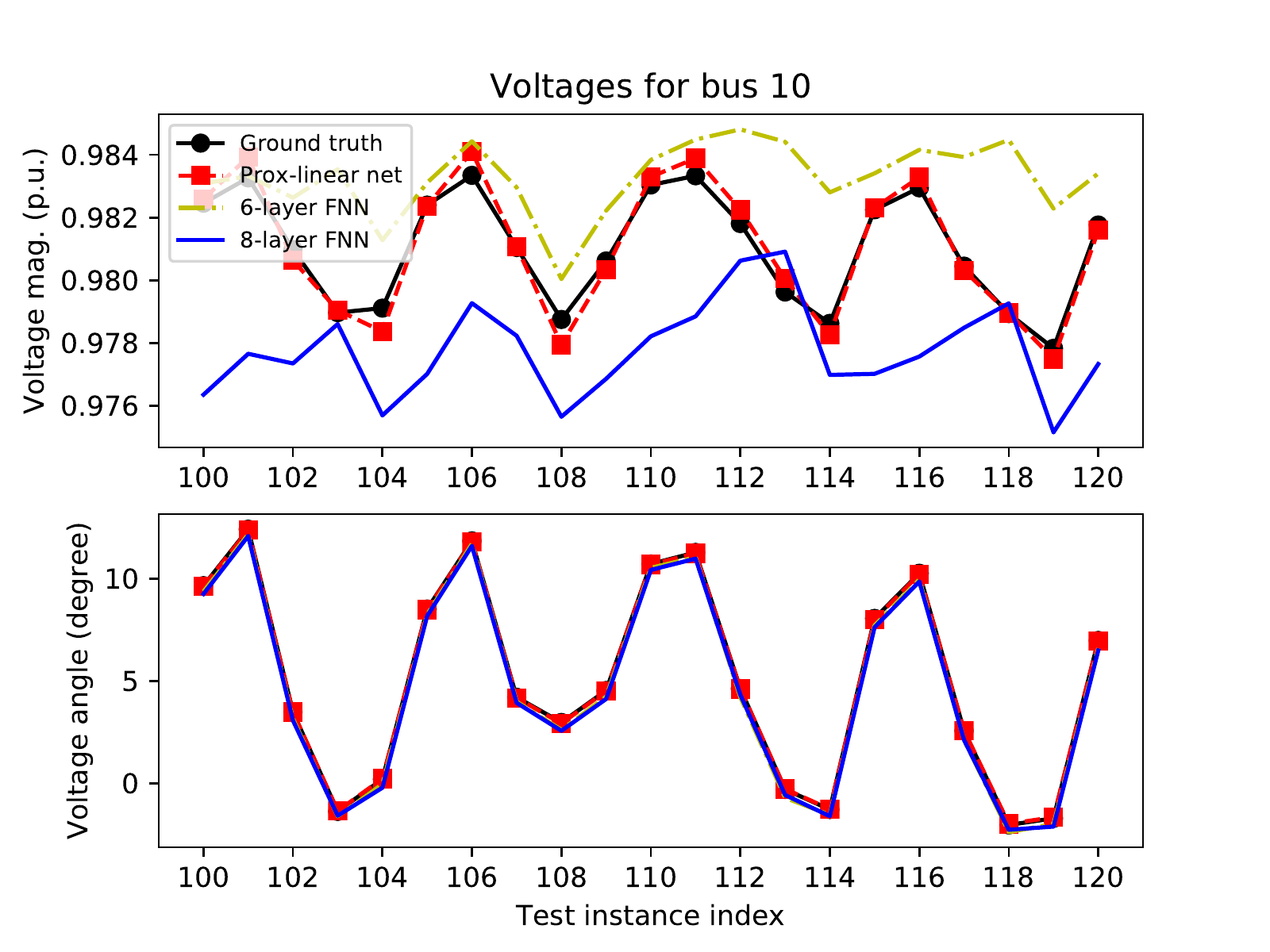}
	\caption{Estimation errors in voltage magnitudes and angles of bus $10$ of the $57$-bus system  from test instances $100$ to $120$.}
	\label{fig:57_bus_10}
\end{figure}

\begin{figure}[t]
	\centering
	\includegraphics[scale=0.58]{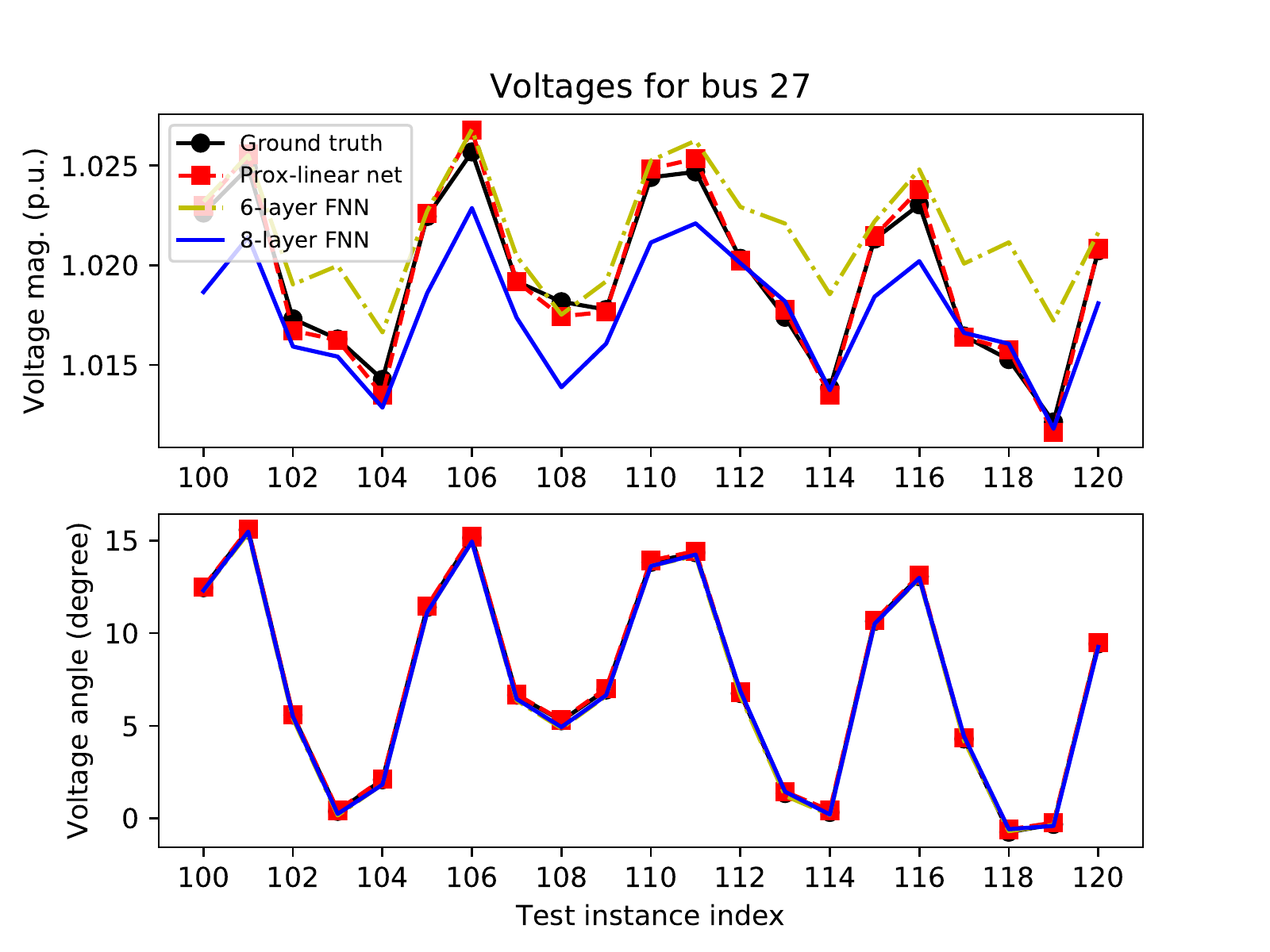}
	\caption{Estimation errors in voltage magnitudes and angles  of bus $27$ of the $57$-bus system from test instances $100$ to $120$.}
	\label{fig:57_bus_27}
\end{figure}

\begin{figure}[t]
	\centering
	\includegraphics[scale=0.58]{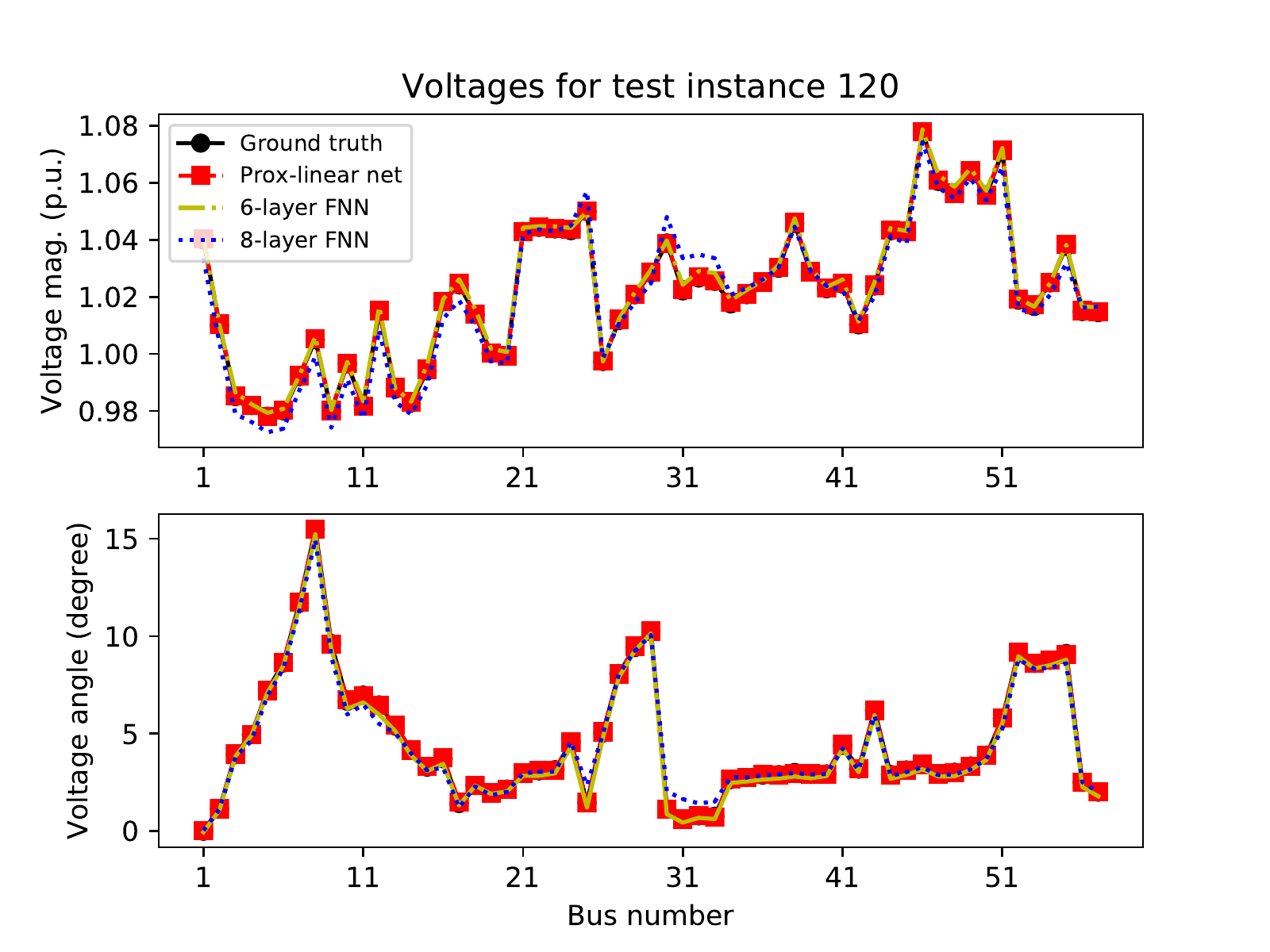}
	\caption{Estimation errors in voltage magnitudes and angles of all the $57$ buses of the $57$-bus system  at test instance $120$.}
	\label{fig:57_test_instance120}
\end{figure}

As far as the computation time is concerned,  the prox-linear net, Gauss-Newton, $6$-layer FNN, and $8$-layer FNN over $1,500$ testing examples are correspondingly $0.0973$s, $14.22$s, $0.0944$s, and $0.0954$s, corroborating the speedup advantage of NN-based PSSE over the traditional Gauss-Newton approach. The ground-truth voltages along with the estimates found by the prox-linear net, $6$-layer FNN, and $8$-layer FNN for bus $10$ and bus $27$ from test instances $100$ to $120$, are shown in Figs. \ref{fig:57_bus_10} and \ref{fig:57_bus_27}, respectively. The true voltages and the estimated ones by NNs for all buses on test instance $120$ are depicted in Fig.~\ref{fig:57_test_instance120}. Evidently, our prox-linear net based PSSE performs the best in all cases. 

The second experiment tests our prox-linear net using the IEEE $118$-bus system, where $18,528$ voltage-measurement pairs were simulated, with $14,822$ pairs employed for training and $3,706$ kept for testing.  The average RMSEs over $3,706$ testing examples for the prox-linear net, Gauss-Newton, $6$-layer FNN and $8$-layer FNN, are $\bf{2.97\times10^{-4}}$, $4.71\times10^{-2}$, $1.645\times10^{-3}$, and $2.366\times10^{-3}$, respectively. Clearly, our prox-linear net yields markedly improved performance over competing alternatives in this case (especially as the system size grows large).  The Gauss-Newton approach performs the worst due to unbalanced grid parameters of this test system. Interestingly, it was frequently observed that the Gauss-Newton iterations minimize the weighted least-squares (WLS) objective function (resulting a loss smaller than $10^{-6}$), but converge to a stationary point that is far away from the simulated ground-truth voltage. This is indeed due to the nonconvexity of the WLS function, for which multiple optimal solutions often exist. Depending critically on initialization, traditional optimization based solvers can unfortunately get stuck at any of those points. In sharp contrast, data-driven NN-based approaches nicely bypass this hurdle. 

\begin{figure}[t]
	\centering
	\includegraphics[scale=0.58]{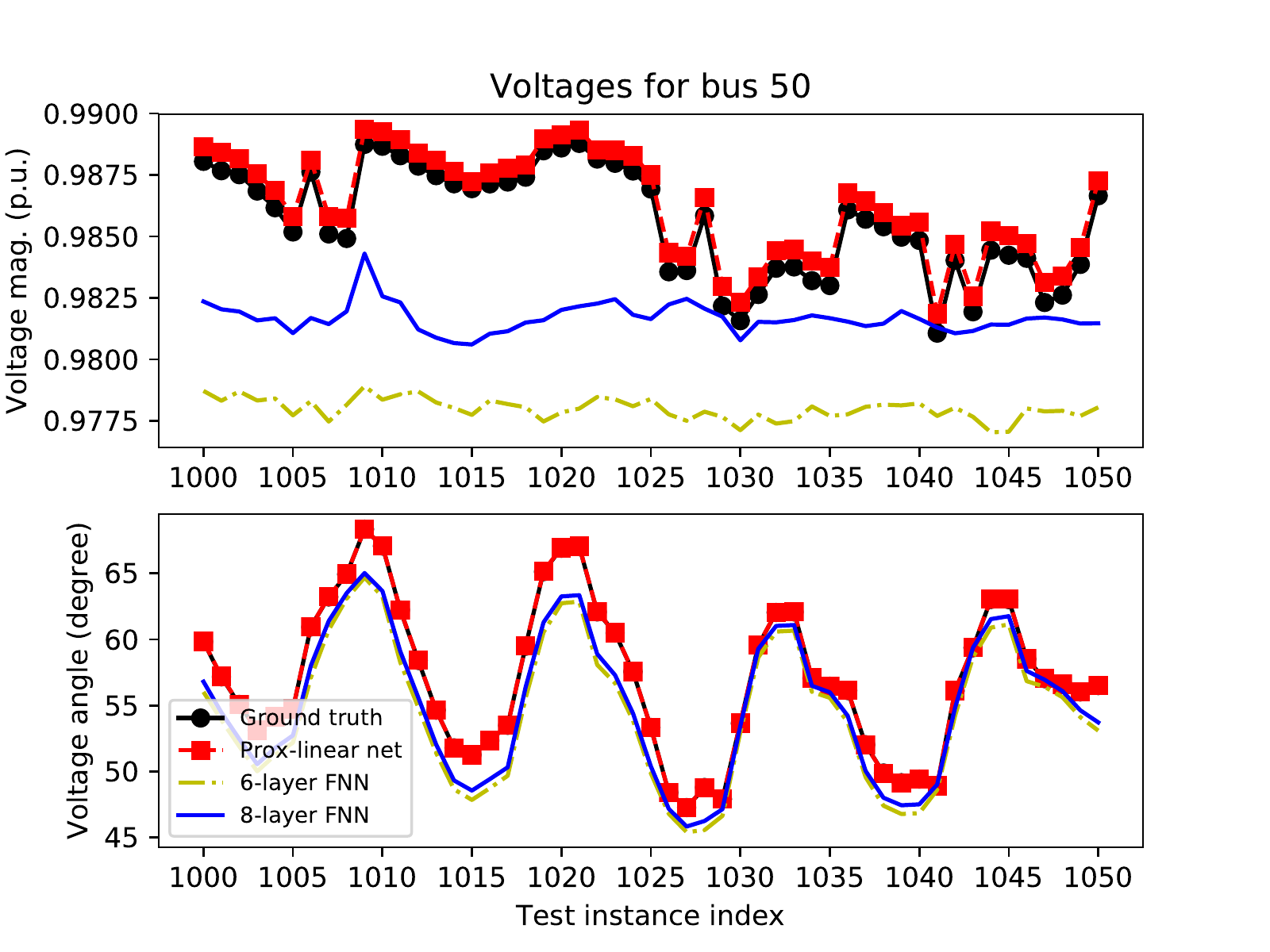}
	\caption{Estimation errors in voltage magnitudes and angles of bus $50$ of the $118$-bus system  from instances $1,000$ to $1,050$.}
	\label{fig:118_bus50}
\end{figure}

\begin{figure}[h!]
	\centering
	\includegraphics[scale=0.58]{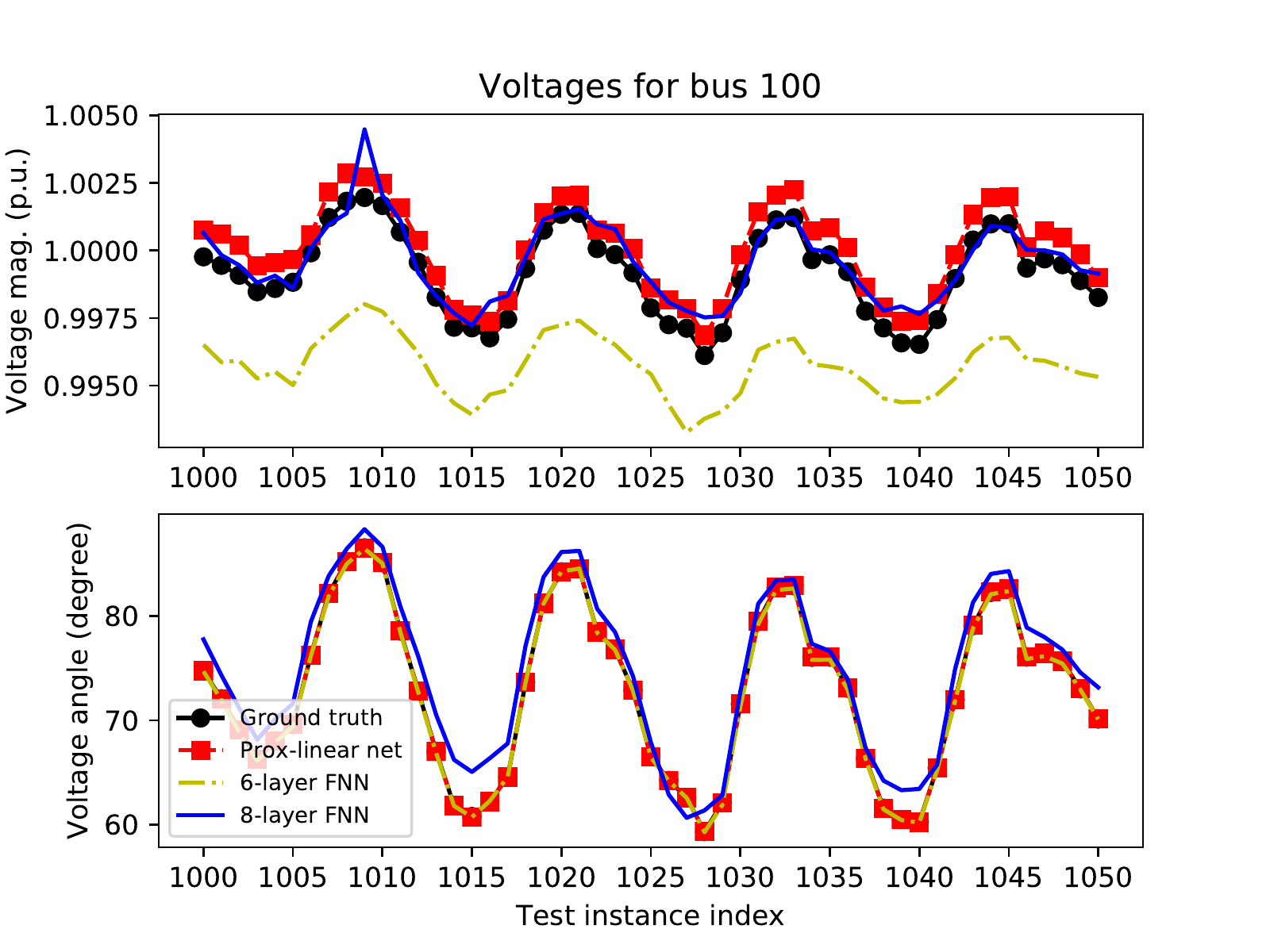}
	\caption{Estimation errors in voltage magnitudes and angles of bus   $100$  of the $118$-bus system from instances $1,000$ to $1,050$.}
	\label{fig:118_bus100}
\end{figure}

\begin{figure}[h!]
	\centering
	\includegraphics[scale=0.58]{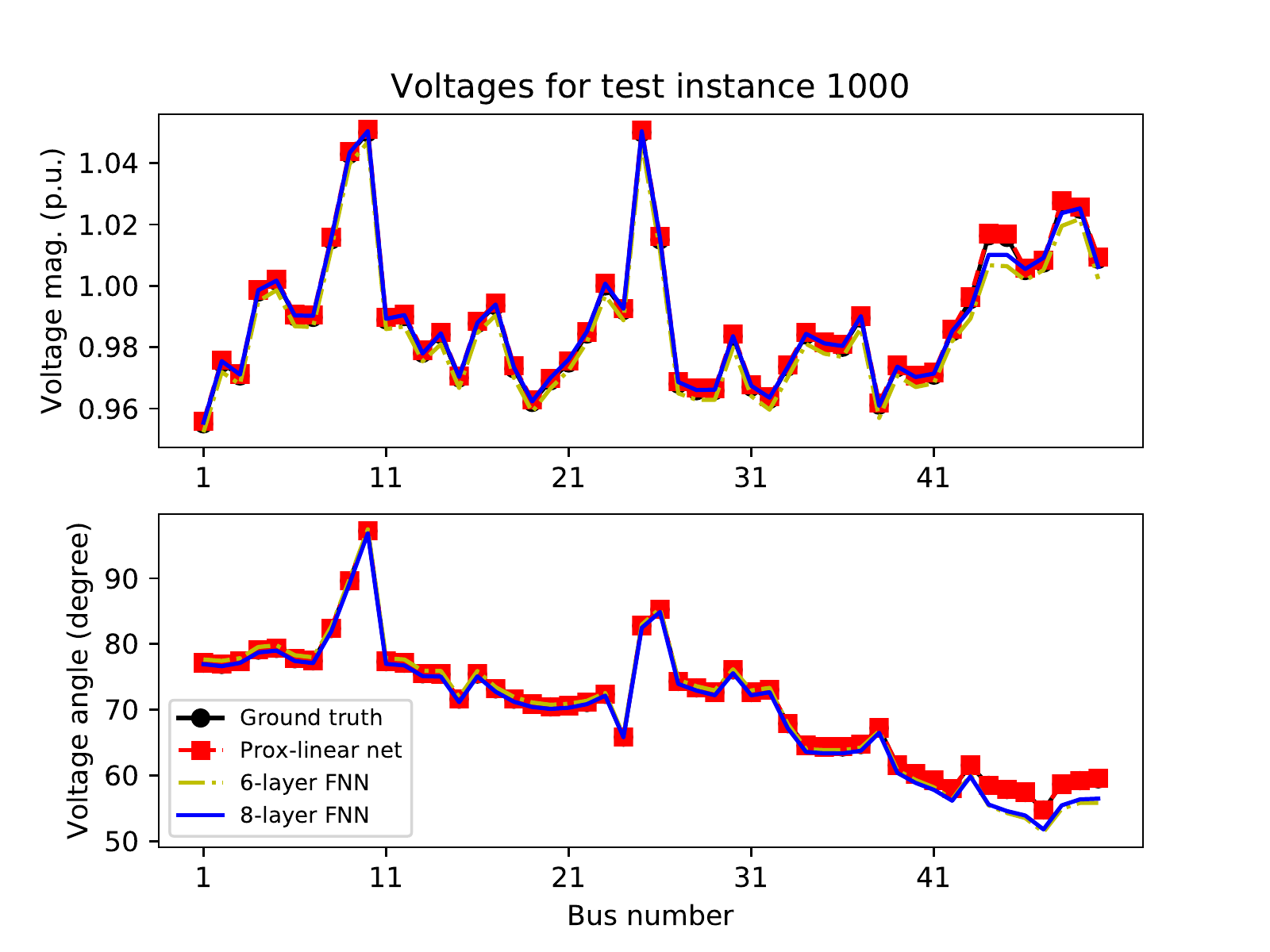}
	\caption{Estimation errors in voltage magnitudes and angles of the first 50 buses of the $118$-bus system  at test instance $1,000$.}
	\label{fig:118_test_instance1000}
	\vspace{-0.6cm}
\end{figure}

In terms of runtime, the prox-linear net, Gauss-Newton, $6$-layer FNN, and $8$-layer FNN, over $3,706$ testing examples are correspondingly $0.3323$s, $183.4$s, $0.2895$s, and $0.3315$s, corroborating again the efficiency of NN-based approaches. The ground-truth voltage along with estimates obtained by the prox-linear net, 6-layer FNN, and 8-layer FNN, for bus $50$ and bus $100$ at test instances $1,000$ to $1,050$, are depicted in Figs. \ref{fig:118_bus50} and \ref{fig:118_bus100}, respectively. In addition, the actual voltages and their estimates for the first fifty buses  on test instance $1,000$ are depicted in Fig.~\ref{fig:118_test_instance1000}. In all cases, our prox-linear net yields markedly improved performance relative to competing alternatives.

\subsection{Deep RNNs for state forecating}

This section examines our RNN based power system state forecasting scheme.
 The forecasting performance was evaluated in terms of the normalized RMSE $\|\check{\mathbf{v}}-\mathbf{v}\|_2 / N$ of the forecast $\check{\mathbf{v}}$ relative to the ground truth $\mathbf{v}$.

Specifically, deep RNNs with $l = 3$, $r=10$, and ReLU activation functions were trained and tested on the ground-truth voltage time series, and on the estimated voltage time series from the prox-linear net. We will refer to the latter as `RNNs with estimated voltages' hereafter. The number of hidden units per layer in RNNs was kept the same as the input dimension, namely $57 \times 2 = 114$ for the $57$-bus system, and $118 \times 2 = 236$ for the $118$-bus system. 
For comparison, a single-hidden-layer FNN ($2$-layer FNN)~\cite{do2009forecasting2}, and a VAR($1$) model \cite{hassanzadeh2016short} based state forecasting approaches were adopted as benchmarks. The average RMSEs over $20$ Monte Carlo runs for the RNN, RNN with estimated voltages, $2$-layer FNN, and VAR($1$) are respectively $2.303 \times 10^{-3}$,  $2.305 \times 10^{-3}$, $3.153 \times 10^{-3}$, and $6.772 \times 10^{-3}$ for the $57$-bus system, as well as $2.588\times10^{-3}$,  $2.751\times10^{-3}$, $4.249\times10^{-3}$, $6.461\times10^{-3}$ for the $118$-bus system. These numbers demonstrate that our deep RNN with estimated voltages offers comparable forecasting performance relative to that with ground-truth voltages. 
Although both FNN and VAR($1$) were trained and tested using ground-truth voltage time-series, they perform even worse than our RNN trained with estimated voltages. 

\begin{figure}[t]
	\centering
	\includegraphics[scale=0.58]{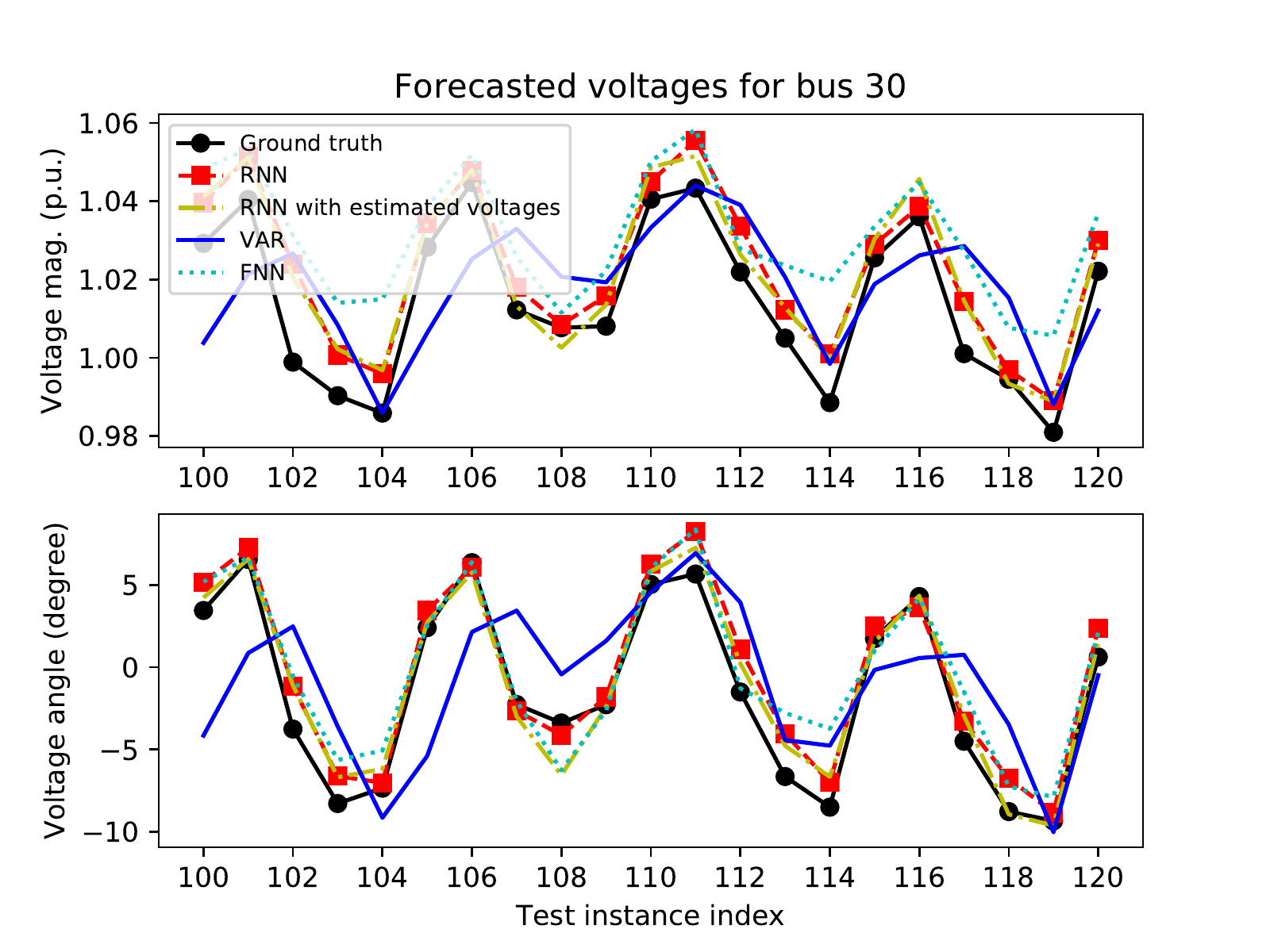}
	\caption{Forecasting errors in voltage magnitudes and angles of bus $30$  of the $57$-bus system from test instances $100$ to $120$.}
	\label{fig:57_FASE_bus30}
\end{figure}

\begin{figure}[t]
	\centering
	\includegraphics[scale=0.58]{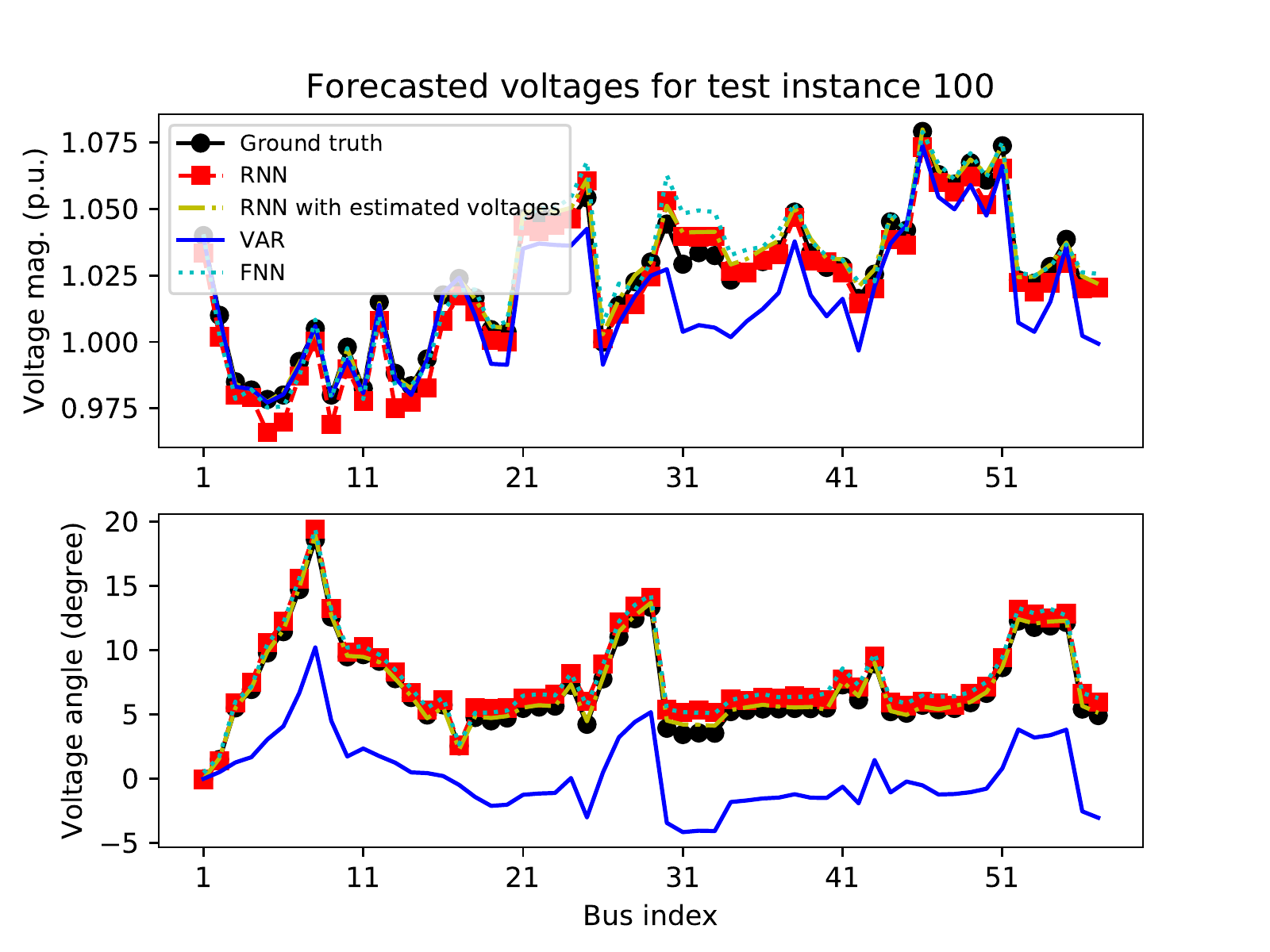}
	\caption{Forecasting errors in voltage magnitudes and angles of all the $57$ buses of the $57$-bus system  at test instance $100$.}
	\label{fig:57_FASE_test_instance120}
    \vspace{-0.3cm}
\end{figure}

\begin{figure}[t]
	\centering
	\includegraphics[scale=0.58]{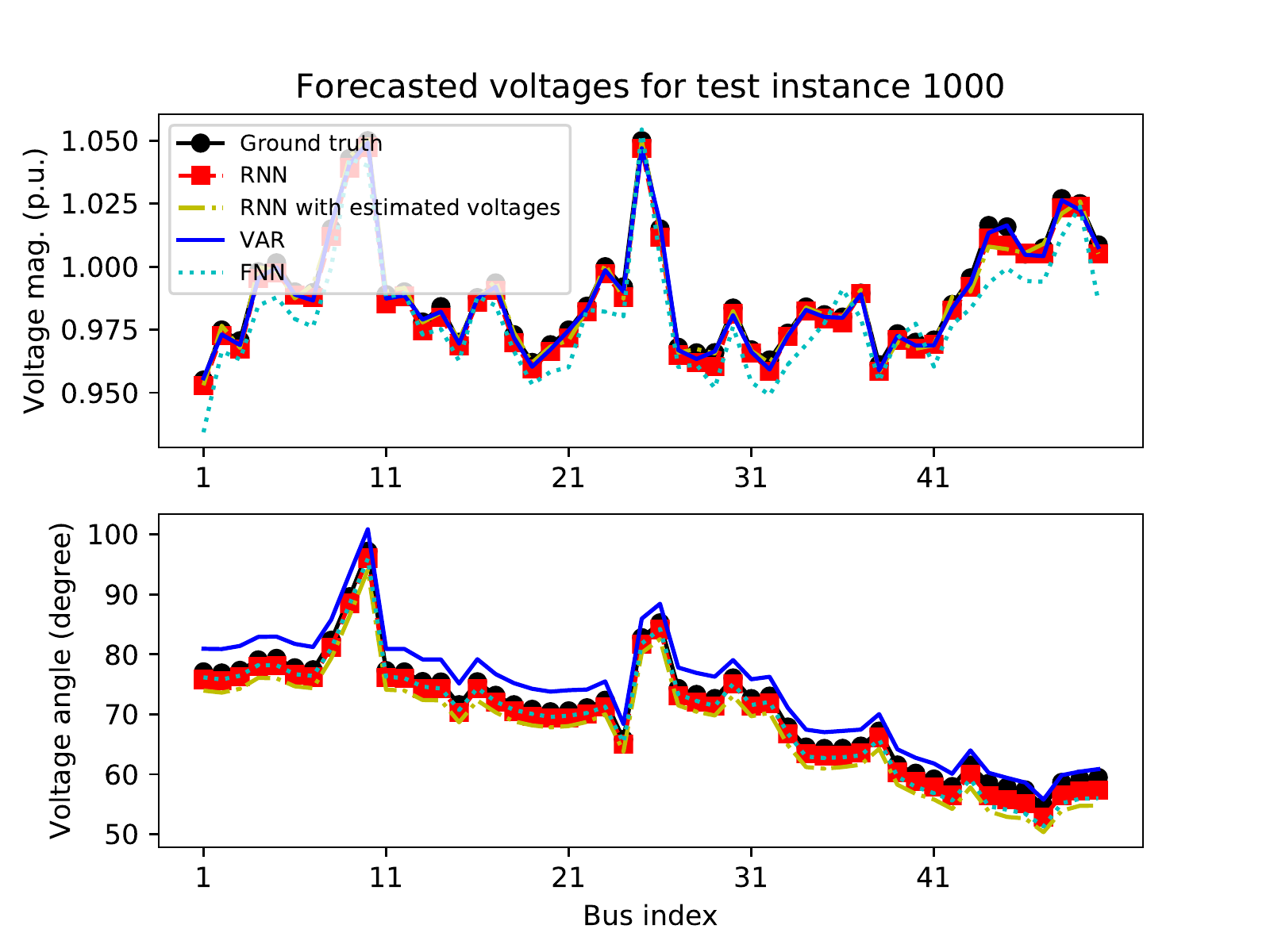}
	\caption{Forecasting errors in voltage magnitudes and angles of the first 50 buses of the $118$-bus system at instance $1,000$.}
	\label{fig:118_FASE_test_instance1000}
	\vspace{-0.3cm}
\end{figure}

The true voltages and their forecasts provided by the deep RNN, RNN with estimated voltages, $2$-layer FNN, and VAR($1$) for bus $30$ of the $57$-bus system from test instances $100$ to $120$, as well as all buses on test instance $100$, are reported in Figs. \ref{fig:57_FASE_bus30} and \ref{fig:57_FASE_test_instance120}, accordingly. The ground-truth and forecast voltages for the first $50$ buses of the $118$-bus system on test instance $1,000$ are depicted in Fig.~\ref{fig:118_FASE_test_instance1000}. Curves illustrate that our deep RNN based approaches perform the best in all cases.
 
\section{Conclusions}\label{sec:conclusions}
This paper dealt with real-time power system monitoring (estimation and forecasting) by building on data-driven DNN advances. Prox-linear nets were developed for PSSE, that combine NNs with traditional physics-based optimization approaches. Deep RNNs were also introduced for power system state forecasting from historical (estimated) voltages. Our model-specific prox-linear net based PSSE is easy-to-train, and computationally inexpensive. The proposed RNN-based forecasting accounts for the  long-term nonlinear dependencies in the voltage time-series, enhances PSSE, and offers situational awareness ahead of time. Numerical tests on the IEEE $57$- and $118$-bus benchmark systems using real load data illustrate the merits of our developed approaches relative to existing alternatives. 

Our current and future research agenda includes specializing the DNN-based estimation and forecasting schemes to distribution networks. Our agenda also includes `on-the-fly' RNN-based algorithms to account for dynamically changing environments, and corresponding time dependencies. 

\bibliographystyle{IEEEtran}
\bibliography{myabrv,power}
\end{document}